**Vine-DB: An RGB image dataset for multi-species ornamental vine segmentation**

**Authors**

Saroj Burlakoti [a*], Utsav Bhandari [b], Aaron Etienne [b], Shital Poudyal [a]

**Affiliations**

[a] Department of Plants, Soils and Climate, Utah State University, Utah, 84322, USA

[b] Department of Applied Sciences, Technology and Education, Utah State University, Utah, 84322, USA

*Saroj.burlakoti@usu.edu

**Abstract**

The Vines-DB dataset contains 1,218 original high-resolution RGB images of seven ornamental vine species collected under field conditions at the Utah Agricultural Experiment Station's Greenville Research Farm in Logan, Utah, USA. The dataset was generated from 168 individual vine plants that were transplanted in 2022 and photographed repeatedly across multiple months during the 2023 and 2024 growing seasons (July–October). Images were captured with an iPhone 16 Pro equipped with a 48 MP camera between 10:00 AM and 12:00 PM under natural daylight. Vines were grown on 4 ft × 8 ft (1.2 m × 2.4 m) trellises and photographed from a distance of 1 m against solid black or white Styrofoam backdrops to improve contrast and reduce background noise. The dataset includes *Akebia quinata, Campsis radicans, Hydrangea anomala petiolaris, Lonicera × heckrottii, Campsis × tagliabuana* 'Madame Galen', *Parthenocissus quinquefolia,* and *Wisteria floribunda*. All original images were manually annotated in Roboflow by trained annotators to produce polygon-based instance segmentation masks for eight classes, including seven species and background. After preprocessing and data augmentation, the working dataset was expanded to 2,307 images for model development and evaluation. The augmented dataset was divided into 2,019 training images, 192 validation images, and 96 test images using stratified sampling to maintain balanced species representation. Vines-DB supports the development and evaluation of deep learning models for multi-class instance segmentation in precision horticulture and urban ecology. The dataset enables applications such as automated canopy cover estimation, species identification, and scalable field phenotyping. In addition, repeated monthly imaging of the same plants captures temporal variation in canopy development and plant appearance, increasing the dataset's utility for segmentation benchmarking under realistic field conditions.

# SPECIFICATIONS TABLE

| | |
|---|---|
| **Subject** | Computer Sciences |
| **Specific subject area** | Multi-class instance segmentation of ornamental vine canopies using deep learning for automated canopy area estimation and phenotyping |

| Type of data | RGB images |
|---|---|
| **Data collection** | Data were collected July–October (2023–2024) at the Utah Agricultural Experiment Station. Images were captured using an iPhone 16 Pro 48 MP camera mounted on a tripod (0.6 m high), positioned 1 m from the canopy between 10 AM and 12 PM. Vines on 4 ft x 8 ft (1.2 m × 2.4 m) trellises were photographed against black/white Styrofoam backdrops for contrast. Exclusion criteria involved a manual review to remove images with motion blur, occlusion, or artifacts. Images were annotated in Roboflow. Normalization included resizing to 640 x 640 pixels via letterbox padding and scaling values to the range [0, 1]. |
| **Data source location** | Utah State University, Logan, Utah, United States of America<br><br>Latitude and longitude from the agricultural field: [41.765945, -111.810036; 41.765933, -111.809509; 41.766379, -111.810019; 41.766292, -111.809487; 41.766368, -111.809620] |
| **Data accessibility** | Repository name: Vines-DB<br><br>Data identification number: **10.17605/OSF.IO/YJHCK**<br><br>Direct URL to data: https://osf.io/yjhck/overview |

# VALUE OF THE DATA

- These data provide a high-resolution, field-based RGB image dataset of seven ornamental vine species captured under realistic environmental conditions, including natural variability in canopy structure, illumination, and seasonal growth. This makes the dataset particularly valuable for developing and benchmarking computer vision models for complex plant architectures.
- The dataset can be reused by researchers in precision horticulture, computer vision, and urban ecology to train, validate, and compare deep learning models for multi-class instance segmentation, species classification, and canopy area estimation tasks.
- The inclusion of polygon-based instance segmentation annotations across eight classes (seven species plus background) enables detailed pixel-level analysis and supports the development of advanced models requiring high-quality labeled data.
- The dataset spans multiple months and growing seasons, allowing researchers to investigate temporal changes in canopy development, phenology, and plant responses under field conditions.
- These data can be used for a range of analytical approaches, including supervised deep learning, transfer learning, and model benchmarking, as well as integration with other datasets to improve model generalizability across environments and species.

# BACKGROUND

Canopy cover area is an important trait in horticulture because it is closely related to irrigation management, crop growth assessment, and plant performance evaluation [1]. Traditional approaches often rely on manual tracing or semi-automated thresholding tools such as ImageJ, which are labor-intensive and difficult to scale for structurally complex species such as ornamental vines [2,3]. Twining growth habits, overlapping foliage, and variable canopy architecture make these species especially challenging for conventional image-based measurements.

The Vines-DB dataset was developed to support the transition from manual image analysis to automated deep learning–based field phenotyping, building on recent advances in agricultural image segmentation [4]. The dataset contains 1,218 original, field-collected, annotated RGB images from 168 unique plant specimens representing seven ornamental vine species. These original images serve as the core dataset, while augmented derivatives were generated separately for model training.

This data article describes the image acquisition protocol, annotation procedure, dataset organization, and preprocessing workflow used to prepare Vines-DB for multi-species ornamental vine segmentation. The dataset is intended to support reproducible development and evaluation of computer vision models for canopy segmentation, species identification, and automated canopy area estimation.

# DATA DESCRIPTION

The Vines-DB repository is organized to support direct use in computer vision workflows while clearly distinguishing between original images and augmented derivatives. The core dataset contains 1,218 original high-resolution RGB images and their corresponding instance segmentation annotations for seven ornamental vine species and one background class. These original images were collected from 168 individual vine plants that were transplanted in 2022 and photographed repeatedly across multiple months during the 2023 and 2024 growing seasons. This repeated imaging captures temporal variation in canopy development, foliage density, and plant appearance under field conditions.

In addition to the original dataset, preprocessing and augmentation were applied for model development, expanding the working dataset to 2,307 images. Augmented images are documented separately from the original non-augmented dataset to ensure transparency in dataset composition and reuse. All images were resized to 640 × 640 pixels to maintain compatibility with standard deep learning pipelines.

Table 1 summarizes the distribution of original annotated images across the seven species included in Vines-DB. Figure 1 presents representative examples of raw RGB images and their corresponding expert-annotated instance segmentation masks.

**Table 1**
Distribution of original annotated images and unique plants across the seven ornamental vine species included in Vines-DB. Image counts refer only to original, non-augmented images. Unique plants imaged refers to the number of individual plants photographed for each species, regardless of repeated imaging across months.

| **Class Name** | **Common Name** | **Original Images** | **Unique Plants Imaged** |
|---|---|---|---|
| *Akebia quinata* ‘Shirobana’ | White-flowered chocolate vine | 209 | 24 |

| *Campsis radicans* ‘Monbal’ | Balboa Sunset trumpet vine | 163 | 24 |
|---|---|---|---|
| *Campsis × tagliabuana* ‘Madame Galen’ | Madam Galen trumpet vine | 170 | 24 |
| *Hydrangea anomala petiolaris* | Climbing hydrangea | 131 | 24 |
| *Lonicera × heckrottii* ‘Goldflame’ | Goldflame honeysuckle | 186 | 24 |
| *Parthenocissus quinquefolia* | Virginia creeper | 162 | 24 |
| *Wisteria floribunda* ‘Texas Purple’ | Texas Purple Japanese Wisteria | 197 | 24 |
| **Total** | | **1,218** | **168** |

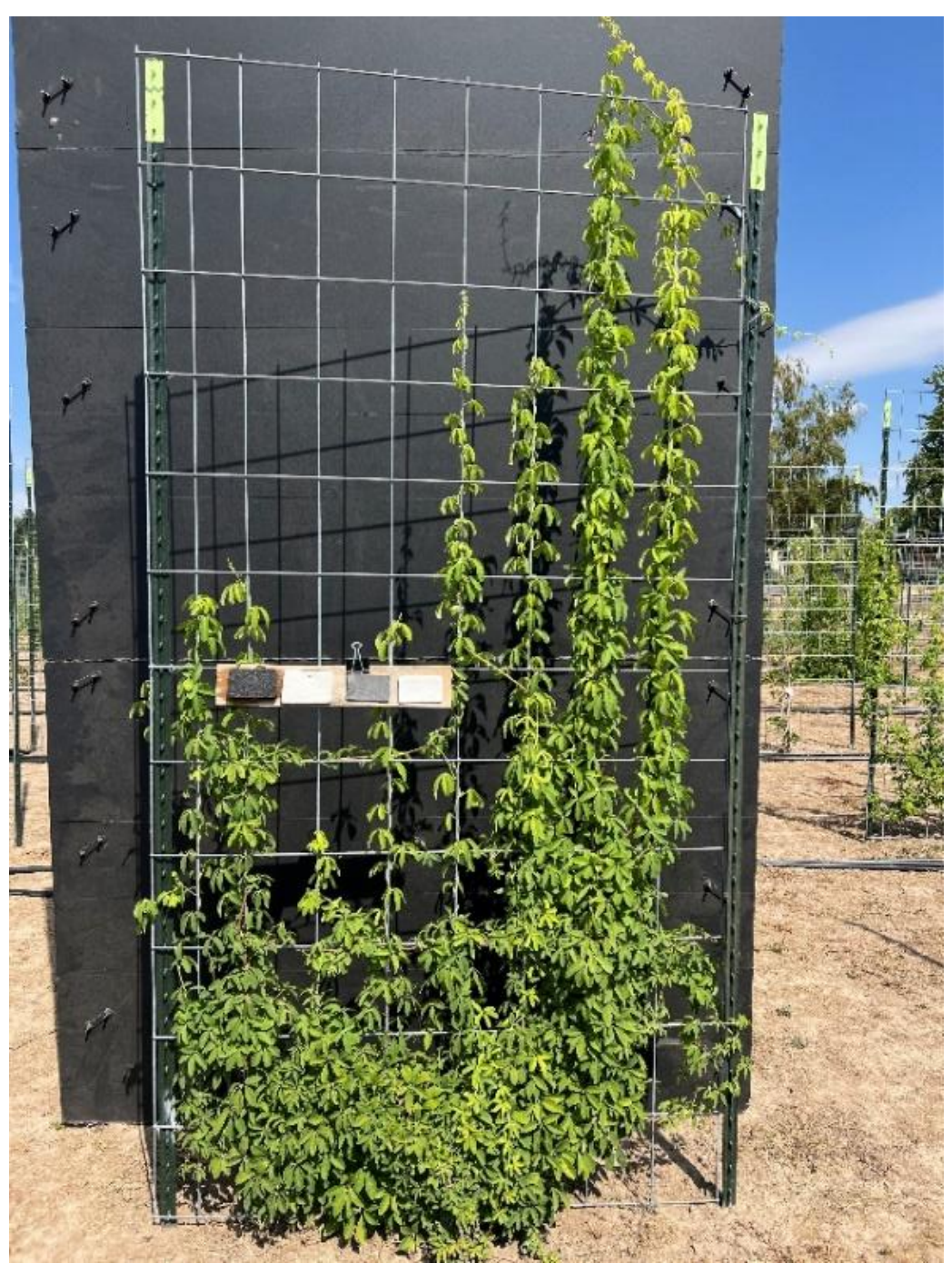

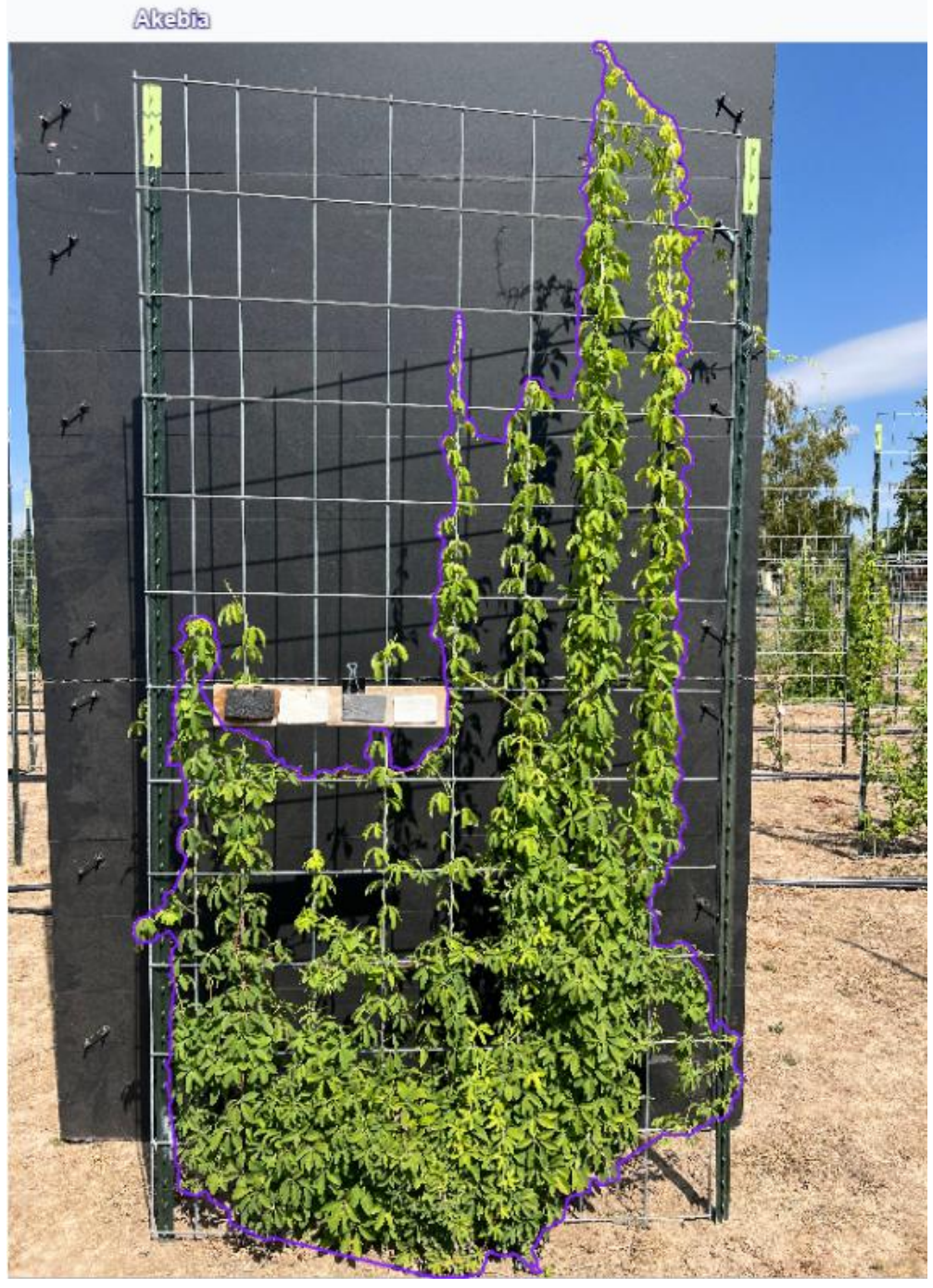

**Figure 1:** *Sample images from the Vine-DB dataset. The left panel shows raw RGB field images, while the right panel shows corresponding expert-annotated instance segmentation masks overlaid on the original images.*

# EXPERIMENTAL DESIGN, MATERIALS AND METHODS

**Study site and experimental setup**

Data were collected at the Utah Agricultural Experiment Station, Greenville Research Farm, Logan, Utah, USA. The study was conducted during the 2023- 2024 growing seasons (July through October).

Seven ornamental vine species (*Akebia quinata*, *Campsis radicans*, *Hydrangea anomala petiolaris*, *Lonicera × heckrottii* 'Goldflame', *Campsis × tagliabuana* 'Madame Galen', *Parthenocissus quinquefolia*, and *Wisteria floribunda*) were grown on trellis systems measuring 4 ft × 8 ft (1.2 m × 2.4 m) [5,6]. Plants were spaced approximately 0.9 m apart within rows. To standardize background conditions, vines were photographed against solid Styrofoam panels (white in the establishment year and black in subsequent collections), each approximately 6 ft × 8 ft (1.8 m × 2.4 m). This setup minimized background variability and improved canopy visibility.

**Image acquisition**

RGB images were captured using an iPhone 16 Pro (Apple Inc., Cupertino, CA, USA) equipped with a 48 MP main camera. The camera was mounted on a tripod at a height of approximately 0.6 m and positioned at 1 m from the canopy. Images were acquired between 10:00 AM and 12:00 PM under consistent daylight conditions to minimize shadow variation. Images were initially recorded in HEIF format and converted to JPEG for compatibility with downstream processing.

**Data preprocessing and augmentation**

All original images were resized to 640 × 640 pixels using letterbox padding to preserve aspect ratio, and pixel values were normalized to the range [0, 1] prior to model input. Data augmentation was applied only for model development using the Roboflow preprocessing pipeline (Roboflow, Inc., Des Moines, IA, USA). Augmentation techniques included random rotation (±15°), horizontal and vertical flipping, brightness and contrast adjustment (±20%), and random cropping (90–110% scaling) [7]. Each original image generated one to two augmented variants depending on species representation and canopy complexity. These augmented images were used to expand the training workflow and are not counted as original images in the core dataset. To ensure transparency and compliance with dataset reporting standards, original images and augmented derivatives are documented and stored separately in the repository. After preprocessing and augmentation, the final working dataset used for model development contained 2,307 images.

**Annotation procedure**

All images were manually annotated using the Roboflow platform. Polygon-based instance segmentation masks were created for each visible plant. Eight classes were defined: background, Akebia, Campsis, Hydrangea, Lonicera, Madame Galen, Parthenocissus, and Wisteria. Two trained annotators independently labeled a subset of 150 images to assess annotation consistency. Inter-annotator agreement was evaluated using mean Intersection over Union (IoU), yielding 0.89 ± 0.06 [8]. Discrepancies were resolved through consensus review.

**Dataset structure and split**

The augmented working dataset used for model development was divided into training (2,019 images), validation (192 images), and test (96 images) subsets using stratified sampling to maintain proportional representation of all species [9]. Images from the same plant were assigned to a single subset to reduce the risk of data leakage between splits. Annotations were exported in COCO instance segmentation format, including image metadata, category IDs, segmentation polygons, and bounding boxes. Optional YOLO-format labels were also generated to support object detection and segmentation workflows.

**Computational framework and software**

All data processing, annotation, and model benchmarking were conducted using the following computational environment:

- Annotation tool: Roboflow (web-based interface)
- Programming language: Python 3.10
- Core libraries: OpenCV for image processing, NumPy for numerical operations, and Matplotlib for visualization
- Deep learning frameworks: PyTorch and Ultralytics for training and evaluating instance segmentation models
- Hardware: Workstation equipped with an NVIDIA GeForce RTX 3090 GPU
- Reference baseline: ImageJ (Version 1.54r) was used for manual canopy area measurements

## LIMITATIONS

The dataset is limited to a single geographic location, which may introduce background and environmental biases. Images represent the active growing season (July–October) and do not include early spring emergence or dormancy stages. Additionally, morphological similarity among species and dense canopy structures may introduce minor inconsistencies in manual annotations.

## ETHICS STATEMENT

The current work does not involve human subjects, animal experiments, or any data collected from social media platforms.

## CRediT AUTHOR STATEMENT

**Saroj Burlakoti:** Conceptualization, Methodology, Software, Data curation, Writing- Original draft preparation, Review and Editing. **Utsav Bhandari**: Methodology, Software, Data Curation, Writing- Review & Editing. **Aaron Etienne**: Resources, Data Curation, Writing- Review & Editing, Supervision. **Shital Poudyal:** Conceptualization, Writing- Reviewing and Editing, Supervision, Project administration, Funding acquisition

## ACKNOWLEDGEMENTS

This work was supported by the USDA National Institute of Food and Agriculture through the Specialty Crop Multi-State Program, administered by the California Department of Food and Agriculture, under the project 'Climate Ready Vines for The Western United States'.

## DECLARATION OF COMPETING INTERESTS

The authors declare that they have no known competing financial interests or personal relationships that could have appeared to influence the work reported in this paper.